\documentclass[runningheads]{llncs}
\usepackage{graphicx}
\usepackage[utf8]{inputenc}
\usepackage[english]{babel}

\begin{document}
\title{Autonomous Vehicle Control: End-to-end Learning in Simulated Urban Environments}
\titlerunning{End-to-end Learning in Simulated Urban Environments}
%
\author{Hege Haavaldsen\thanks{These authors contributed equally to this work.} \orcidID{0000-0001-6372-1989} \and
Max Aasbø\textsuperscript{*}\orcidID{0000-0003-3462-165X} \and Frank Lindseth\thanks{Corresponding author.}\orcidID{0000-0002-4979-9218}}

\authorrunning{H. Haavaldsen, M. Aasbø, F. Lindseth}

\institute{Norwegian University of Science and Technology, Trondheim, Norway}
\maketitle              
\begin{abstract}

In recent years, considerable progress has been made towards a vehicle's ability to operate autonomously. An end-to-end approach attempts to achieve autonomous driving using a single, comprehensive software component. Recent breakthroughs in deep learning have significantly increased end-to-end systems' capabilities, and such systems are now considered a possible alternative to the current state-of-the-art solutions.

This paper examines end-to-end learning for autonomous vehicles in simulated urban environments containing other vehicles, traffic lights, and speed limits. Furthermore, the paper explores end-to-end systems' ability to execute navigational commands and examines whether improved performance can be achieved by utilizing temporal dependencies between subsequent visual cues.

Two end-to-end architectures are proposed: a traditional Convolutional Neural Network and an extended design combining a Convolutional Neural Network with a recurrent layer. The models are trained using expert driving data from a simulated urban setting, and are evaluated by their driving performance in an unseen simulated environment. 

The results of this paper indicate that end-to-end systems can operate autonomously in simple urban environments. Moreover, it is found that the exploitation of temporal information in subsequent images enhances a system's ability to judge movement and distance.

\end{abstract}
\section{Introduction}
\label{section:introduction}

We are currently at the brink of a new paradigm in human travel: the fully autonomous, self-driving car. Only 50 years ago, cars were completely analog devices with almost no mechanisms for assisting the driver. Over the decades, additional features, controls, and technologies have been integrated, and cars have evolved into exceedingly complex machines.

In recent years, substantial progress has been made towards a vehicle's ability to operate autonomously. Primarily, two different approaches have emerged. The prevailing state of the art approach is to divide the problem into a number of sub-problems and solve them by combining techniques from computer vision, sensor fusion, localization, control theory, and path planning. This approach requires expert knowledge in several domains and often results in complex solutions, consisting of several cooperating modules.

Another approach is to develop an end-to-end solution, solving the problem using a single, comprehensive component, e.g., a deep neural network. A technique for training such a system is to employ imitation learning. This entails studying expert decisions in different scenarios, to find a mapping between the perceived environments and the executed actions. While some believe that the black-box characteristics of such systems makes them untrustworthy and unreliable, others point to recent years' advances in deep-learning and argue that end-to-end solutions show great potential.

However, end-to-end systems cannot make the correct navigational decision solely based on a perceived environment. It is also necessary to incorporate a user's intent in situations that require a decision (e.g., when approaching an intersection). Hence, an end-to-end system should be able to receive and adapt to navigational commands.

The objective of this paper is to investigate end-to-end systems' ability to drive autonomously in simulated urban environments. Specifically, to study their performance in environments containing other vehicles, traffic lights, and speed limits; to examine their ability to oblige navigational commands in intersections; and to explore if a system can improve its performance by utilizing temporal dependencies between subsequent visual cues.

This paper seeks to combine different aspects from recent research in the field of end-to-end learning for autonomous vehicles. Concretely, the use of navigational commands as network input, and the exploitation of temporal dependencies between subsequent images. There have been no attempts - to our knowledge - to combine both techniques in one system. Hopefully, this can lead to a more complete end-to-end system and improved driving quality.

The rest of this paper is organized as follows. Section \ref{section:related-work} presents previous related work, while Section \ref{section:environment} addresses the environment in which the data was collected and the experiments were conducted. Section \ref{section:data-generation} reviews the collection and preprocessing of the data. The model architectures are presented in Section \ref{section:model-architectures}. Section \ref{section:experimental-setup} and \ref{section:experimental-results} covers the experimental setup and results, while section \ref{section:discussion} discusses the results. Finally, Section \ref{section:conclusion} covers the conclusions.

\section{Related Work}
\label{section:related-work}

There have been several advances in end-to-end learning for autonomous vehicles over the last decades. The first approach was seen already in 1989 when a fully connected neural network was used to control a vehicle \cite{ref:alvinn}. In 2003 a proof-of-concept project called DAVE emerged \cite{ref:dave}, showing a radio controlled vehicle being able to drive around in a junk-filled alley and avoiding obstacles. DAVE truly showed the potential of an end-to-end approach. Three years later NVIDIA developed DAVE-2 \cite{ref:dave2}, a framework with the objective to make real vehicles drive reliably on public roads. DAVE-2 is the basis for most end-to-end approaches seen today \cite{ref:conditional-e2e,ref:add-nav}. The project used a CNN to predict a vehicle’s steering commands. Their model was able to operate on roads with or without lane markings and other vehicles, as well as parking lots and unpaved roads.

Codevilla et al. \cite{ref:conditional-e2e} further explored NVIDIA's architecture by adding navigational commands to incorporate the drivers intent into the system and predicted both steering angle and acceleration. The authors proposed two network architectures: a \textit{branched network} and a \textit{command input network}. The \textit{branched network} used the navigational input as a switch between a CNN and three fully connected networks, each specialized to a single intersection action, while the \textit{command input network} concatenated the navigational command with the output of the CNN, connected to a single fully connected network.

Hubschneider et al. \cite{ref:add-nav} proposed using turn signals as control commands to incorporate the steering commands into the network. Furthermore, they proposed a modified network architecture to improve driving accuracy. They used a CNN that receives an image and a turn indicator as input such that the model could be controlled in real time. To handle sharp turns and obstacles along the road the authors proposed using images recorded several meters back to obtain a spatial history of the environment. Images captured 4 and 8 meters behind the current position were added as an input to make up for the limited vision from a single centered camera. 

Eraqi,  H.M. et. al.\cite{ref:temporal-dependencies} tried to utilize the temporal dependencies by combining a CNN with a Long Short-Term Memory Neural Network. Their results showed that the C-LSTM improved the angle prediction accuracy by 35 \% and stability by 87 \%.

\section{Environment}
\label{section:environment}

Training and testing models for autonomous driving in the physical world can be expensive, impractical, and potentially dangerous. Gathering a sufficient amount of training data requires both human resources and suitable hardware, and it can be time consuming to capture, organize and label the desired driving scenarios. Moreover, the cost of unexpected behavior while testing a model may be considerable.

An alternative is to train and test models in a simulated environment. A simulator can effectively provide a variety of corner cases needed for training, validation, and testing; while removing safety risks and material costs. Additionally, the labeling of the dataset can be automated, removing the cost of manual labeling, as well as the potential of human error.

The drawback, however, is the loss of realism. A simulation is only an imitation of a real-world system, and a model trained on only simulated data may not be able to function reliably in the real world. Nonetheless, a simulator can give a good indication of a model's actual driving performance and serves well for benchmarking different models. Once a model can perform reliably in a simulated environment, the model can be fine-tuned for further testing in real environments.

In this paper, the CARLA simulator \cite{ref:carla-paper} is used to gather training data and to evaluate the proposed models. CARLA is an open source simulator built for autonomous driving research and provides an urban driving environment populated with buildings, vehicles, pedestrians, and intersections. 

\section{Data Generation}
\label{section:data-generation}
\subsection{Data Collection}
\label{section:data-collection}

When performing imitation learning, the quality of the training data plays a significant role in a model's ability to perform reliably in different conditions. However, a model trained only using expert data in ideal environments may not learn how to recover from perturbations. To overcome this, several types of driving data was captured. Expert driving was captured using CARLA's built-in autopilot, resulting in center-of-lane driving while following speed-limits. To capture more volatile data, a randomly generated noise value was added to the autopilot's outgoing control signal. This resulted in sudden shifts in the vehicle's trajectory and speed, which the autopilot subsequently tried to correct. To eliminate undesirable behavior in the training set, only the autopilot's response to the noise was collected, not the noisy control signal. Finally, recovery from possible disaster states was captured by manually steering the vehicle into undesired locations, e.g., the opposite lane, or the sidewalk; while recording the recovery.

For each recorded frame, images from three forward-facing cameras (positioned at the left, center, and right side of the vehicle) were captured, along with the vehicle's control signal (i.e., steering angle, throttle, and brake values), and additional information (i.e., speed, speed limit, traffic light state, and \textit{High-Level Command}). The \textit{High-Level Command} (HLC) is the active navigational command, labeling the data with the user's current intent. Possible HLCs are: \textit{follow lane}, \textit{turn left at the next intersection}, \textit{turn right at the next intersection}, and \textit{continue straight ahead at the next intersection}.

Two different datasets were gathered, one for training and one for testing. The training set was captured in CARLA's Town 1, while the test set was captured in Town 2. Data were gathered in four different weather conditions: Clear noon, cloudy noon, clear sunset, and cloudy sunset. The training set contained driving data captured both with and without other vehicles. The test set exclusively contained driving data alongside other vehicles. All data were captured in environments without pedestrians. Table \ref{table:datasets} summarizes the gathered datasets. 
All expert data was captured driving 10 km/h below the speed limit to match the velocity of other vehicles. 

\begin{table}
    \caption {The collected datasets. An observation contains the captured data from a single rendered frame in the simulator.}
    \label{table:datasets}
    \begin{tabular}{| l | c | c |}
    \hline
    Dataset & Number of Observations & Size [GB]\\
    \hline
    Training & 117 889 & 31.5\\
    Testing & 23 173 & 8.32\\
    \hline
    \end{tabular}
\end{table}

\subsection{Data Preparation}
\label{section:data-preparation}

For each recorded observation, a data sample was created containing the center image, the vehicle's control signal, and the additional information. Moreover, to simulate the recovery from drifting out of the lane, two new data samples were generated using the observation's left and right images. To counteract the left and right images' positional offset, the associated steering angle was shifted by +0.1 and -0.1 respectively. 

For each data sample, a new augmented sample was generated using one of the desirable transformations picked at random. These included a random change in brightness or contrast, the addition of Gaussian noise or blur, and the addition of randomly generated dark polygons differing in position and shape. 

When recording the datasets, the majority of the observations were captured driving straight. To prevent an unbalanced dataset, data samples with a small steering angle were downsampled (by removal), while data samples with a large steering angle were upsampled (by duplication). Additionally, the data samples corresponding to the different intersection decisions (i.e., turn left, turn right, or straight ahead) were balanced by analyzing the distributions of HLC-properties in the dataset and downsampling the over-represented choices. Finally, some of the data samples where the vehicle was not moving (e.g., waiting for a red light) were downsampled.

\section{Model Architectures}
\label{section:model-architectures}

In this paper, two related end-to-end architectures are proposed. The first is a Convolutional Neural Network (CNN) inspired by NVIDIA's DAVE-2 \cite{ref:dave2} system, while the second extends the CNN with Long-Short-Term-Memory (LSTM) units to capture temporal dynamic behavior. 

\subsection{CNN Model}
\label{section:cnn-model}

The CNN model consists of two connected modules: a feature extractor and a prediction module. The former uses a CNN to extract useful features from the input image, while the latter combines the detected features with the additional inputs (i.e., current speed, speed limit, traffic light state, and HLC) to predict a control signal (i.e., steering angle, throttle and brake values).

The convolutional part of the model is inspired by the architecture used in NVIDIA's DAVE-2 system \cite{ref:dave2}. The modified network takes a 180x300x3 image as input, followed by a cropping layer and a normalization layer. The cropping layer removes the top 70 pixels from the image, while the normalization layer scales the pixel values between -0.5 and 0.5. Next follows six convolutional layers, all using a ReLU activation function. The first three layers use a 5x5 filter, while the last three use a 3x3 filter. The first four layers use a stride of 2, while the last two use a stride of 1. The output of the last convolutional layer is flattened resulting in a one-dimensional feature layer containing 768 nodes.

\begin{figure}
    \includegraphics[width=\textwidth]{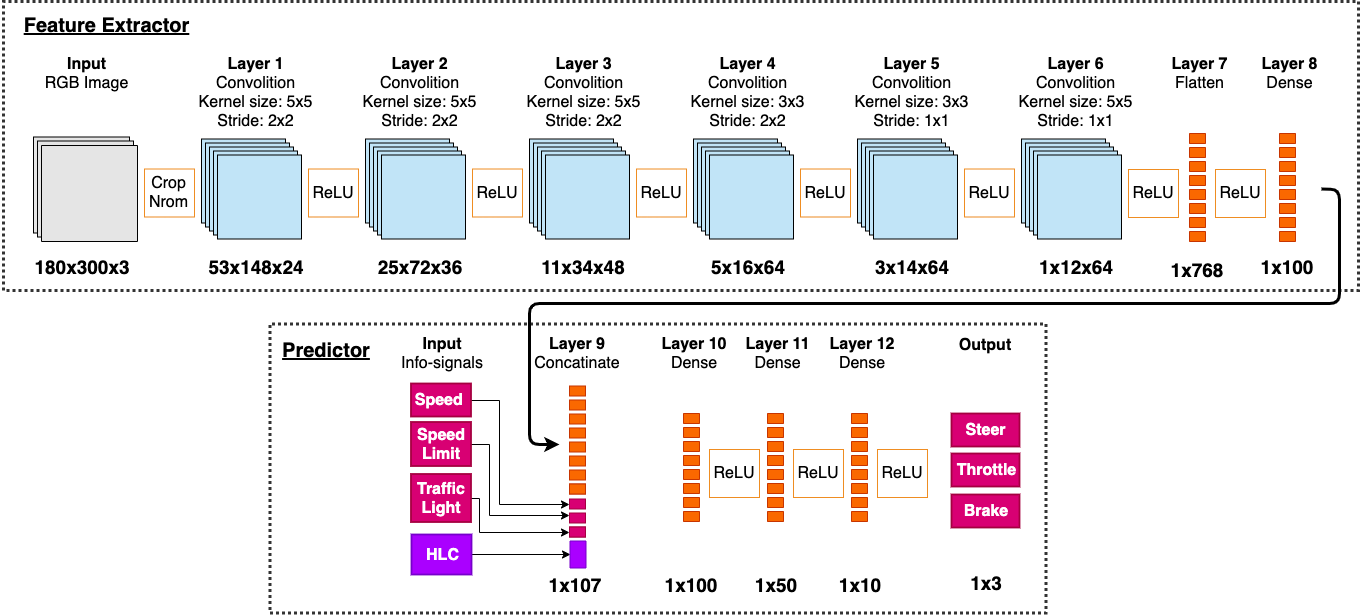}
    \caption{The architecture of the CNN model. The model accepts a single RGB image and predicts a control signal. }  
    \label{figure:cnn-architecture}
\end{figure}

The output from the convolutional layers is concatenated with the additional input containing the speed, speed-limit, traffic-sign, and HLC values. The concatenated layer serves as an input for the predictive part of the model which consists of three dense layers containing 100, 50, 10 nodes respectively. All the dense layers use a ReLU activation function. The last of the dense layers are finally connected to an output layer, consisting of 3 nodes. The complete architecture is shown in Figure \ref{figure:cnn-architecture}.

\subsection{CNN-LSTM Model}

The CNN-LSTM model consists of two connected modules: a feature extractor and a temporal prediction module. The former follows the same architecture as the previous model, shown in Figure \ref{figure:cnn-architecture}. The feature extractor is connected to an LSTM layer with 5 hidden states. The model uses a sequence of feature extractions over time to predict a control signal. This allows the model to learn temporal dependencies between time steps.

\begin{figure}
    \includegraphics[width=1\textwidth]{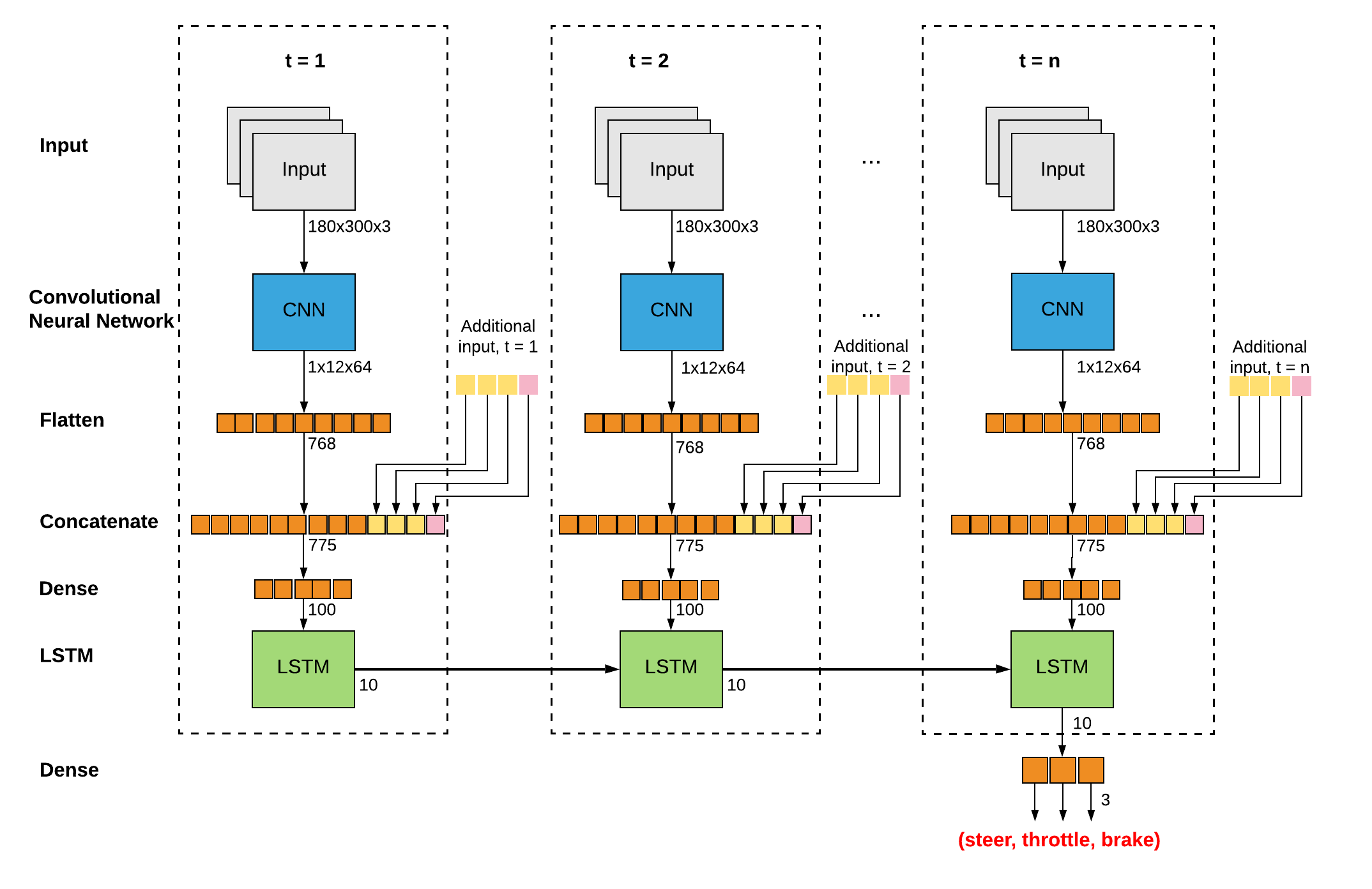}
    \caption{The architecture of the CNN-LSTM model.}
    \label{figure:cnn-lstm-architecture}
\end{figure}

For each time step, the output of the feature extractor is concatenated with the additional input containing speed, speed limit, traffic light, and an HLC. This is sent through a dense layer containing 100 nodes and fed into an LSTM layer with 10 nodes. For each time step in the sequence, the LSTM layer sends its output to itself. At the last time step, the output is sent through a dense layer, consisting of three nodes. This is the final prediction. The complete architecture is shown in Figure \ref{figure:cnn-lstm-architecture}.

\section{Experimental Setup}
\label{section:experimental-setup}

\subsection{Training}

\subsubsection{Training and validation}
The dataset was split into a training set (70\%) and a validation set (30\%). 

\begin{figure}
    \includegraphics[width=1\textwidth]{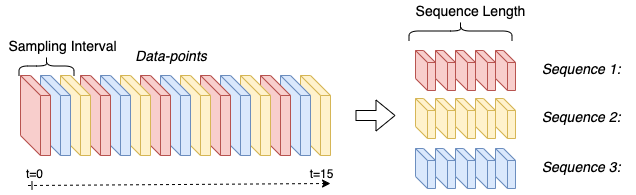}
    \caption{The structuring of sequences from an array of data samples. A sequence length of five and a sampling interval of three is used.}
    \label{figure:lstm-data-struct}
\end{figure}

For the CNN-LSTM, the data samples were further structured into sequences of length five, using a sampling interval of three. The sequence length determines the number of time steps the LSTM layer is able to remember, while the sampling interval decides the period between successive individual time steps within the sequences. Figure \ref{figure:lstm-data-struct} illustrates the structuring of sequences from 15 data samples.

\subsubsection{Hyperparameters}

Both models were trained using an Adam optimizer \cite{ref:adam} and an Mean Squared Error loss function. The models were trained for a 100 epochs, with a batch size of 32 data-samples. The models' weights were recorded after each epoch along with the associated validation error. After the training was complete, the weights associated with the lowest validation error were chosen for further testing.

\subsection{Testing}

After training, each model's predictive and real-time performance was measured. The predictive performance was tested by exposing the models to the unseen test set while calculating the average prediction error. The real-time performance was tested by letting the models control a simulated vehicle in CARLA's second town. Each model was to drive through a predefined route ten times. The test evaluator provided HLCs to the model before each intersection. A model's performance was measured in the number of route completions, the average route completion percentage, and the number of failures. Model failures were recorded according to severity. Touching a lane line was considered a minor failure, while a low-speed rear-ending or an object collision was considered a moderate failure. Object collisions without recovery or an ignored HLC were considered a severe failure. Catastrophic failures consisted of either entering the opposite lane, disregarding a red traffic light, or colliding with oncoming traffic.

\section{Experimental Results}
\label{section:experimental-results}

\subsection{Validation and Test Error}
After 100 epochs of training, both models' training and validation loss had stabilized. The CNN and CNN-LSTM had their lowest validation error after epoch 66 and 81 respectively. The models were then evaluated on the unseen test set from CARLA's second town. On the test set, the CNN was able to predict a control signal with an average error of 0.023, a 43\% increase compared to its validation error. The CNN-LSTM was able to predict a control signal with an average error of 0.022, a 57\% increase compared to its validation error.

\begin{table}
    \caption {Test and validation loss during the training of the models.}
    \label{table:val_test_loss}
    \begin{tabular}{|l |c| c|}
    \hline
    Model & Validation Loss & Test Loss \\
    \hline
    CNN & 0.016 & 0.023 \\
    CNN-LSTM & 0.014 & 0.022 \\
    \hline
    \end{tabular}
\end{table}

\subsection{Real-time Test in Simulated Environment}
\label{section:result-experiment}
Both models were real-time tested in CARLA's second town. To test their ability to handle various driving scenarios in an urban environment, each model drove a predefined route ten times. The results are described below. Two videos were created to show a successful run and some common failures. One demonstrates the CNN model \cite{ref:cnn-model} and the other demonstrates the CNN-LSTM model \cite{ref:cnn-lstm-model}.

\begin{table}[ht]
    \caption { Summary of test results. Each model attempted to drive a predefined route (Figure \ref{figure:town2_map}) ten times. A model's performance were measured in the number of route completions and the average route completion percentage. }
    \label{table:results_completion1}
    \begin{tabular}{| l | c | c |}
    \hline
    Model & Route Completions & Avg. Route Completion \\
    \hline    
    CNN  & 2 &  $56 \%\pm 39$  \\
    CNN-LSTM  & 5 &  $81 \%\pm 9$ \\
    \hline
    \end{tabular}
\end{table}

\begin{table}[ht]
    \caption {Average failures per run. The models' failures were recorded according to severity. Touching a lane line was considered a minor failure, while a low-speed rear-ending or an object collision was considered a moderate failure. Object collisions without recovery or an ignored HLC were considered a severe failure. A catastrophic failure consisted of either entering the opposite lane, disregarding a red traffic light, or colliding with oncoming traffic.}
    \label{table:results_completion2}
    \begin{tabular}{|l | c | c | c | c |}
    \hline
    Model & Minor & Moderate & Severe & Catastrophic \\
     \hline
    CNN & $2.10 \pm 1.73$ &  $0.65 \pm 0.99$ &  $0.25 \pm 0.44$ &  $0.23 \pm 0.57$ \\
    CNN-LSTM & $2.90 \pm 1.72$ & $0.10 \pm 0.31$ & $0.05 \pm 0.22$ & $0.17 \pm 0.38$ \\
    \hline
    \end{tabular}
\end{table}

\subsubsection{CNN}

Out of ten runs, the CNN model was able to complete the predefined route twice, with an average completion percentage of 56\% over all runs. The model performed well on lane following and drove reliably in the center of the lane most of the time. The model handled most intersections but had a tendency to perform very sharp right turns. This resulted in 2.1 minor failures (i.e., the vehicle touching the lane line) per run. It always tried to follow the provided HLC, and never ignored a traffic light. It was able to handle complex light conditions, such as direct sunlight and dark shadows in the road. The model hit objects without recovery several times, leading to 0.5 severe failures per run. The vehicle rarely hit objects outside of the lane, but occasionally struggled to stop for other vehicles, leading to several low speed rear-end collisions. In total, the model had 0.1 object collisions, and 1.2 rear-end collisions per run. Finally, 0.7 times per run it struggled to find the lane after a turn, leading to a catastrophic failure. The model usually held the speed limit but found it hard to slow down fast enough to a speed limit when the speed was high. The results are summarized in Table \ref{table:results_completion1} and Table \ref{table:results_completion2}.
 
\subsubsection{CNN-LSTM}
The CNN-LSTM model was able to complete the predefined route five out of ten times, with an average completion percentage of 81\% over all runs. It drove reliably in the center of the lane most of the time but tended to perform sharp turns. This lead to 2.9 minor failures (i.e., lane line touches) per run. The model always tried to follow the provided HLC, and never ignored a traffic light.  It managed to handle various light conditions, such as direct sunlight and dark shadows in the road. The vehicle rarely struggled to stop for other objects or vehicles, resulting in 0.1 object collisions and 0.1 rear-end collisions per run. Finally, 0.5 times per run it struggled to find the lane after a turn, leading to a catastrophic failure. The model adapted well to the different speed limits. The results are summarized in Table \ref{table:results_completion1} and Table \ref{table:results_completion2}.

\section{Discussion}
\label{section:discussion}

This paper proposed two architectures for an end-to-end system: a traditional CNN inspired by NVIDIA's DAVE-2 system \cite{ref:dave2}, and an extended design combining the CNN with an LSTM layer to facilitate learning of temporal relationships. Both models were able to follow a lane consistently and reliably. Any disturbances or shifts in the trajectory were quickly corrected, without being overly sensitive. The CNN exhibited less volatile steering compared to the CNN-LSTM in the high-speed stretch of the route, but the CNN-LSTM outperformed the CNN in turns following high-speed stretches. Additionally,  both models obeyed all traffic lights and always tried to follow the provided HLC at intersections. 

When introducing other vehicles, the differences between the models became more apparent. The CNN-LSTM adapted its speed according to traffic, and only rear-ended another vehicle once throughout the whole experiment. In scenarios where another vehicle blocked most of the view (e.g., when following another vehicle closely in a turn), the CNN-LSTM seemed to be able to use past predictions as a guide. The CNN, on the other hand, experienced more trouble when driving alongside other vehicles. Although the model, to some degree, adapted its speed according to traffic, it often failed to react upon sudden changes. This led to frequent rear-endings throughout the experiment. 

The difference between the models' performance may be explained in their architectural differences. The CNN-LSTM model used five subsequent observations when making predictions. This allowed it, by all indication, to learn some important temporal dependencies - acquiring some knowledge about the relationships between movement, change in object size, and distance. The CNN model, however, interpreted each observation independently, which restrained its ability to understand motion. It still learned to brake when approaching a vehicle, but was not able to differentiate between fast approaching and slow approaching objects. Predictions related to distance were solely dependent on the size of objects. Moreover, the CNN could not rely on past predictions when faced with confusing input, which seemed to result in more unreliable behavior. 

It should be mentioned that although the CNN model had 28 \% less minor failures than the CNN-LSTM model, its completion rate was 32 \% lower than the CNN-LSTM.  The reduction in minor failures was probably a result of the lower average route completion, not an indication of better performance.

\subsection{Consistency with Related Work}
The implemented models in this paper are based on the architecture in \cite{ref:dave2}. The authors were able to use a CNN to drive on trafficked roads with and without lane markings, parking lots and unpaved roads. This complies with this paper's results. Even though the implemented models were not tested on unmarked roads or parking lost, they were able to drive on roads with lane marking, both on roads with and without pavements. 

Codevilla et. al. \cite{ref:conditional-e2e} claimed that their \textit{command input network} performed inadequately when executing navigational commands. This does not comply with the results of this paper. The proposed architecture takes the navigational command as input after the CNN, in a similar matter to the \textit{command input network}, but was able to execute the given navigational commands with a high degree of success. 

In \cite{ref:add-nav} the turn indicators of the car was used as the navigational commands, which were sent as input to the network. The authors did not use an RNN, but fed three subsequent images to three CNNs and concatenated the output. It was able to perform lane following, avoid obstacles and change lanes. The navigational commands in this paper were introduced to the network in a similar way, and both approaches were able to execute the navigational commands. The proposed system was not tested for lane changes, but seeing achievements in similar approaches indicates that this should be possible. 

The CNN model in this paper was extended with an LSTM to utilize temporal dependencies. A similar approach was attempted in \cite{ref:temporal-dependencies}. They showed that adding temporal dependencies improved both the accuracy and stability of a model using a single CNN. Similar results can be seen in this paper. 

\section{Conclusion}
\label{section:conclusion}

The results of the experiments indicates that end-to-end systems are able to operate autonomously in simulated urban environments. The proposed systems managed to follow lanes reliably in varying lighting conditions and were not disrupted by disturbances or shifts in trajectory. They were able to abide by traffic lights and speed limits and learned to execute different navigational commands at intersections.

Both systems managed to adapt its speed according to traffic, but their ability to respond to sudden changes varied. The CNN-LSTM were, by all indication, able to acquire some insight into the relationships between movement, distance, and change in the perceived size of objects. The regular CNN, interpreting each observation independently, was not able to learn these essential temporal dependencies. Hence, the results suggest that exploiting temporal information in subsequent images improves an end-to-end systems ability to drive reliably in an urban environment. 

Even though the systems' performed several mistakes during testing, their achievements demonstrated great potential for using end-to-end systems to accomplish fully autonomous driving in urban environments. 

%
%
%
\bibliographystyle{splncs04}
\bibliography{references}

\begin{thebibliography}{10}
\providecommand{\url}[1]{\texttt{#1}}
\providecommand{\urlprefix}{URL }
\providecommand{\doi}[1]{https://doi.org/#1}

\bibitem{ref:dave2}
Bojarski, M., Testa, D.D., Dworakowski, D., Firner, B., Flepp, B., Goyal, P.,
  Jackel, L.D., Monfort, M., Muller, U., Zhang, J., Zhang, X., Zhao, J., Zieba,
  K.: End to end learning for self-driving cars. CoRR  \textbf{abs/1604.07316}
  (2016), \url{http://arxiv.org/abs/1604.07316}

\bibitem{ref:conditional-e2e}
Codevilla, F., M{\"{u}}ller, M., Dosovitskiy, A., L{\'{o}}pez, A., Koltun, V.:
  End-to-end driving via conditional imitation learning. CoRR
  \textbf{abs/1710.02410} (2017), \url{http://arxiv.org/abs/1710.02410}

\bibitem{ref:carla-paper}
Dosovitskiy, A., Ros, G., Codevilla, F., L{\'{o}}pez, A., Koltun, V.: {CARLA:}
  an open urban driving simulator. CoRR  \textbf{abs/1711.03938} (2017),
  \url{http://arxiv.org/abs/1711.03938}

\bibitem{ref:temporal-dependencies}
Eraqi, H.M., Moustafa, M.N., Honer, J.: End-to-end deep learning for steering
  autonomous vehicles considering temporal dependencies. CoRR
  \textbf{abs/1710.03804} (2017), \url{http://arxiv.org/abs/1710.03804}

\bibitem{ref:cnn-model}
Haavaldsen, H., Aasbø, M.: Autonomous driving using a cnn: An example of
  end-to-end learning, \url{https://youtu.be/Q37jTFZjK2s}

\bibitem{ref:cnn-lstm-model}
Haavaldsen, H., Aasbø, M.: Autonomous driving using a cnn-lstm: An example of
  end-to-end learning, \url{https://youtu.be/ADzEHGmIDQQ}

\bibitem{ref:add-nav}
{Hubschneider}, C., {Bauer}, A., {Weber}, M., {Zöllner}, J.M.: Adding
  navigation to the equation: Turning decisions for end-to-end vehicle control.
  In: 2017 IEEE 20th International Conference on Intelligent Transportation
  Systems (ITSC). pp.~1--8 (Oct 2017). \doi{10.1109/ITSC.2017.8317923}

\bibitem{ref:adam}
Kingma, D.P., Ba, J.: Adam: {A} method for stochastic optimization. In: 3rd
  International Conference on Learning Representations, {ICLR} 2015, San Diego,
  CA, USA, May 7-9, 2015, Conference Track Proceedings (2015),
  \url{http://arxiv.org/abs/1412.6980}

\bibitem{ref:dave}
Lecun, Y., Cosatto, E., Ben, J., Muller, U., Flepp, B.: Dave: Autonomous
  off-road vehicle control using end-to-end learning. Tech. Rep. DARPA-IPTO
  Final Report, Courant Institute/CBLL, http://www.cs.nyu.edu/\~{
  }yann/research/dave/index.html (2004)

\bibitem{ref:alvinn}
Pomerleau, D.A.: Advances in neural information processing systems 1. chap.
  ALVINN: An Autonomous Land Vehicle in a Neural Network, pp. 305--313. Morgan
  Kaufmann Publishers Inc., San Francisco, CA, USA (1989),
  \url{http://dl.acm.org/citation.cfm?id=89851.89891}

\end{thebibliography}
%










\end{document}